\documentclass[]{fairmeta}
\usepackage{amssymb,amsthm,amsmath}
\usepackage{mathtools}
 \usepackage{xspace}

\renewcommand{\phi}{\varphi}

\usepackage{algorithm}
\usepackage{algorithmic}
\usepackage{float}
\newcommand{\update}{\textsc{update}}

\title{GaLore 2: Large-Scale LLM Pre-Training by Gradient Low-Rank Projection}

\author[1]{DiJia Su}
\author[2]{Andrew Gu}
\author[2]{Jane Xu}
\author[1]{Yuandong Tian}
\author[1]{Jiawei Zhao}

\affiliation[1]{FAIR at Meta AI}
\affiliation[2]{PyTorch}

\abstract{
        Large language models (LLMs) have revolutionized natural language understanding and generation but face significant memory bottlenecks during training. GaLore, Gradient Low-Rank Projection, addresses this issue by leveraging the inherent low-rank structure of weight gradients, enabling substantial memory savings without sacrificing performance. 
        Recent works have further extended GaLore in various aspects, including low-bit quantization and higher-order tensor structures. 
        However, there are several remaining challenges for GaLore, such as the computational overhead of SVD for subspace updates and the integration with state-of-the-art training parallelization strategies (e.g., FSDP).
        In this paper, we present GaLore 2, an efficient and scalable GaLore framework that addresses these challenges and incorporates recent advancements.
        In addition, we demonstrate the scalability of GaLore 2 by pre-training Llama 7B from scratch using up to 500 billion training tokens, highlighting its potential impact on real LLM pre-training scenarios.
    }

\date{\today}

\begin{document}

\maketitle 

\section{Introduction}

Recent advancements in large language models (LLMs) have led to significant breakthroughs in natural language understanding, contextual generation, and complex reasoning. However, the highly overparameterized nature of LLMs presents considerable computational challenges, particularly during pre-training and fine-tuning. Particularly, the memory consumption becomes a significant bottleneck, with substantial memory required for storing model parameters, gradients, and optimizer states. For instance, pre-training a Llama 7B model requires at least 58 GB of memory for just a single batch, making LLM training inaccessible to general AI community, especially without access to high-end hardware.

To tackle these challenges, GaLore, Gradient Low-Rank Projection, has emerged as a promising solution for reducing memory requirements without sacrificing model performance \citep{zhao2024galore}. By leveraging the natural low-rank structure of gradients during training, GaLore projects gradients onto a lower-dimensional subspace, significantly reducing memory costs for storing gradients and optimizer states (such as first and second order estimates in Adam optimizer) while preserving the quality of training. GaLore has significantly reduced memory consumption in both pre-training and fine-tuning. Especially, for the first time, it enables pre-training of a Llama 7B model on a single NVIDIA RTX 4090 GPU with 24GB of memory.

Since its introduction, GaLore has inspired several recent works aimed for enhancing its capabilities. Among them, Q-GaLore combines low-bit weight quantization with low-bit projection to further reduce its memory consumption \citep{zhang2024q}. Tensor-GaLore extends the low-rank projection principle to higher-order tensor structures \citep{george2024tensorgalore}. These developments show the growing interest in using low-rank natures of weight gradients for memory-efficient training.

Despite its broad usage, GaLore faces several challenges. The computational overhead of Singular Value Decomposition (SVD), used for updating gradient subspace projection, remains a bottleneck, particularly for  models with large matrices such as Llama 7B. Additionally, integrating GaLore with state-of-the-art training parallelization strategies, like Fully Sharded Data Parallel (FSDP), still remains an area open for exploration, limiting its scalability for large-scale distributed training.

Another critical question for GaLore is its scalability in real-world pre-training scenarios, where the number of training tokens ranges from 50 billion to over 1 trillion—significantly exceeding the 20 billion tokens evaluated in prior studies. Understanding how GaLore performs under such large-scale settings, with increased computational and memory demands, remains an open challenge.

In this technical report, we present GaLore 2, an enhanced version of GaLore that addresses remaining challenges and incorporates recent advancements. Specifically, GaLore 2 supports fast randomized SVD for subspace updates and integrates with FSDP. We also demonstrate its scalability by training Llama 7B from scratch using up to 500 billion training tokens, highlighting its potential impact on real LLM pre-training scenarios.

\section{Related Works}
Machine learning and Large Language Models (LLMs) have gained much traction due to their incredible ability to solve a variety of tasks \citep{ll1, ll2, ll3, ds1,liu2024spinquantllmquantizationlearned, ds2,ds3, ww1,  su2022narrowing, cohen2025spectral,  su2021musbo, zhou2025sweet, zhou2025gsm, lin2025step, wang2310guiding,  su2020conqur, wu2024meta, liu2024spinquant, su2025token, paulus2024advprompter}. Despite such success, training LLMs requires a substantial memory footprint to accommodate weights, activations, gradients, and optimization states. Efforts to reduce this memory cost include memory-efficient optimization algorithms, quantization, and low-rank adaptation methods. Techniques such as Adafactor \citep{shazeerAdafactorAdaptiveLearning} and 8-bit optimizers \citep{dettmers8bitOptimizersBlockwise2021} lower the memory requirements of gradient statistics and optimizer states. AdaLomo \citep{lvAdaLomoLowmemoryOptimization2023} further reduces memory overhead by fusing backward operations with optimizer updates, eliminating the need to store weight gradients.

Low-Rank Adaptation (LoRA) \citep{hu2021lora} is a widely adopted approach that enables memory-efficient fine-tuning by introducing trainable low-rank adapters while keeping the base model frozen. Building upon LoRA's success, numerous works have extended its capabilities in different directions. For instance, ReLoRA \citep{lialin2023relora} adapts LoRA for pre-training by periodically merging the low-rank updates into the base model, though it requires full-rank warmup to match baseline performance. MultiLoRA \citep{wang2023multilora} enhances LoRA's multi-task learning capabilities by introducing task-specific adapters that can be efficiently composed. Recent works have also taken alternative approaches to reduce memory. For example, BAdam \citep{luo2024badam} reduces memory through parameter partitioning and efficient state management, while LISA \citep{pan2024lisa} employs layer-wise importance sampling to selectively update parameters during training.

\citet{zhao2024galore} propose Gradient Low-Rank Projection (GaLore) to project gradients onto a low-rank subspace to reduce memory consumption. It has been widely used and integrated into various training frameworks, including PyTorch and Hugging Face. 
Recent works further extend GaLore from various aspects \citep{naturalgalore, welore}.
\citet{zhang2024q} develop Q-GaLore to combine low-bit quantization with low-rank projection to further reduce the memory consumption of GaLore.
\citet{george2024tensorgalore} extend the low-rank projection principle to higher-order tensor structures for solving partial differential equations using neural networks.
\citet{liang2024memoryefficientllmtrainingonline} provide comprehensive analysis on the convergence of online subspace descent methods.
\citet{Robert2024LDAdamAO} adopt a moment calibration technique to calibrate the gradient statistics during subspace updates.

\section{Background: Gradient Low-Rank Projection (GaLore)}

As proposed in \citet{zhao2024galore}, GaLore projects gradients onto a low-rank subspace to reduce memory consumption. For example, when applying GaLore to a layer weight matrix $W_t \in \mathbb{R}^{m \times n}$ with $m \leq n$ at iteration $t$, GaLore projects its gradient $G_t \in \mathbb{R}^{m \times n}$ onto a subspace $R_t \in \mathbb{R}^{n \times r}$ using a projection matrix $P_t \in \mathbb{R}^{n \times r}$:
\begin{align*}
    G_t &\gets - \nabla_W \phi_t(W_t), \\
    R_t &= P_t^\top G_t.
\end{align*}
The low-rank subspace gradient $R_t$ is used for gradient accumulation and serves as the input for preconditioned optimizers. In the case of the Adam optimizer, Adam creates two low-rank moments $M_t \in \mathbb{R}^{n \times r}$ and $V_t \in \mathbb{R}^{n \times r}$ to track the statistics and produce a low-rank update $N_t \in \mathbb{R}^{n \times r}$:
\begin{align*}
    M_t &\gets \beta_1 \cdot M_{t-1} + (1 - \beta_1) \cdot R_t, \\
    V_t &\gets \beta_2 \cdot V_{t-1} + (1 - \beta_2) \cdot R_t^2, \\
    N_t &\gets M_t / (\sqrt{V_t} + \epsilon).
\end{align*}

Finally, GaLore reprojects the low-rank update $N_t$ back to the original space using the projection matrix $P_t$ and uses it to update the model weights:
\begin{align*}
    \tilde G_t &\gets \alpha \cdot P_t N_t, \\
    W_t &\gets W_{t-1} + \eta \cdot \tilde G_t,
\end{align*}
where $\alpha$ is a scale factor and $\eta$ is the learning rate. One of the most important part of GaLore is the choice of the projection matrix $P_t$, which affects the next gradient subspace to be optimized. GaLore updates the projection matrix $P_t$ that best matches the spectrum of the current gradient by using Singular Value Decomposition (SVD), such that:
\begin{align*}
    U, S, V &= \text{SVD}(G_t), \\
    P_t &= U[:, :r] \quad \text{if } m \leq n, \\
    P_t &= V[:, :r]^\top \quad \text{if } m > n,
\end{align*}
where the subspace update happens every $T$ steps during training. The complete algorithm is provided in the appendix.

GaLore reduces optimizer memory from $2mn$ to $2nr$ for the Adam optimizer, but it also introduces additional memory costs for storing the projection matrix $P_t$, which costs $mr$. In total, GaLore requires $(mn + mr + 2nr)$ memory, while LoRA requires $(mn + 3mr + 3nr)$ memory. We omit the memory cost for storing weight gradients as it depends on the specific implementation choice in practice. For example, if no gradient accumulation is employed, we can directly adopt per-layer weight update to eliminate this type of memory cost \citep{zhao2024galore}. On the other hand, an additional $nr$ memory cost is introduced for storing the accumulated low-rank gradient $R_t$.

Although GaLore demonstrates significant effectiveness in reducing memory requirements during training, several key challenges and open questions remain unresolved, such as the mechanics behind subspace updates, integration with the latest training parallelization strategies, and scalability for real pre-training scenarios.

\section{GaLore 2}

We present GaLore 2, an efficient and scalable GaLore framework. It includes our analysis and improvements on subspace update, incorporation of related GaLore extensions, and integration with PyTorch FSDP.

\subsection{Subspace Update}
Updating the gradient subspace is a core component of GaLore. 
If we stay too long within one subspace, the parameters are likely to overfit to the subspace and stop decreasing in loss. Therefore, it is necessary to frequently update the subspace to ensure the entire parameter space can be well explored. By default, after a fixed amount of steps, GaLore updates the projection matrix to match the spectrum of the currently received gradient. Specifically, GaLore performs Singular Value Decomposition (SVD) on the gradient matrix and uses either left or right singular vectors as the components of the new projection matrix. 

\begin{figure}[t]
    \centering
    \includegraphics[width=\textwidth]{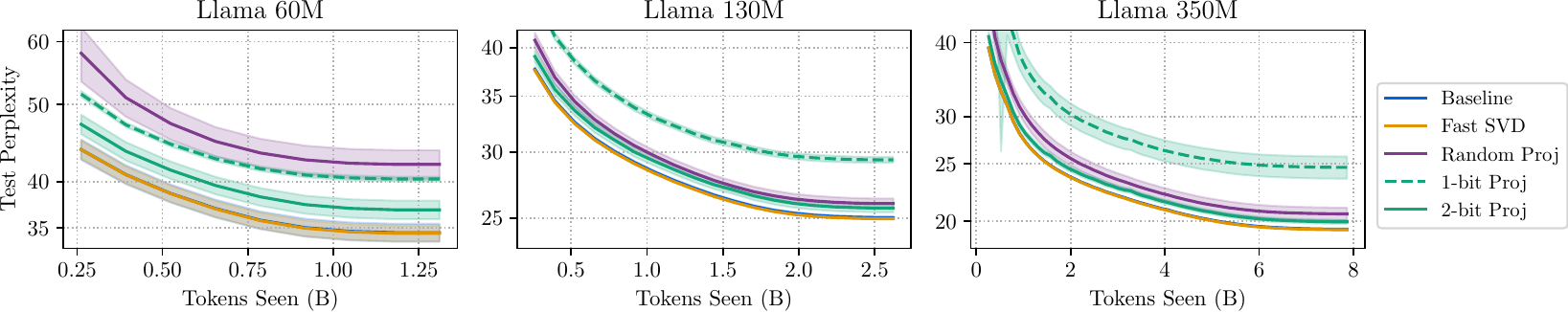} % Adjust the width as needed
    \captionsetup{justification=centering} % Center the caption
    \caption{Comparison of different projection methods across various Llama models.}
    \label{fig:proj_compare}
\end{figure}

\subsubsection{Projection Types}
Although matching the gradient spectrum is a good strategy for finding the next subspace, recent works \citep{zhang2024q} indicate that the projection matrix does not need to exactly match the current gradient spectrum. A quantized and approximated projection matrix can also achieve similar performance. However, we empirically find that the performance degrades when the approximation gap is large. To the extreme, when using a random projection matrix, the performance degrades significantly. It is necessary to carefully choose the trade-off between spectrum approximation and potential memory savings for the projection matrix.
As shown in Figure~\ref{fig:proj_compare}, we compare different projection methods across various models. Random and extremely quantized projection methods significantly degrade performance.

\subsubsection{Fast Randomized SVD}
Computing SVD can be extremely expensive for large matrices. For example, when applying GaLore to the Llama 7B model, the SVD computation takes up to 20 minutes per subspace update. To address this, GaLore 2 adopts the fast randomized SVD method proposed in \cite{halko2011finding}, which achieves a truncated matrix decomposition through randomness. Our empirical results indicate that fast randomized SVD can be 15X faster than the original SVD operation with no loss in accuracy. 
As shown in Figure~\ref{fig:proj_compare}, we compare different projection methods across various models. Figure~\ref{fig:proj_compare} suggests that the fast randomized SVD method fully matches the GaLore baseline across Llama models.

\subsubsection{Randomization and Sign Indeterminacy Issue}
SVD suffers from the sign indeterminacy issue, where its output is not unique, and the sign of the singular vectors can be permuted\footnote{Some SVD implementations provide options to solve sign ambiguity by fixing the signs of ordered loadings for each component \citep{tensorly,scikitlearn}.}. In addition, the randomized SVD algorithm introduces additional randomness to the SVD output. These factors make the projection matrix inconsistent during training. For example, assuming two consecutive gradients are similar to each other, the two SVD operations can generate two sets of singular vectors with opposite signs. This makes frequent subspace updates in GaLore unstable. However, for moderate frequencies of subspace updates (such as 200-500 steps adopted by \citet{zhao2024galore}), we find this issue negligible as the gradients at consecutive subspace updates are different enough.

\subsection{GaLore Extensions}
GaLore 2 also incorporates recent works that enhance GaLore from various aspects, including low-bit quantization and higher-order tensor structures.
\citet{zhang2024q} proposed Q-GaLore to combine low-bit quantization with low-rank projection to further reduce the memory consumption of GaLore.
\citet{george2024tensorgalore} proposed Tensor-GaLore to extend the low-rank projection principle to higher-order tensor structures.
GaLore 2 includes support for both low-bit projection matrices and higher-order tensor structures.
In addition, GaLore 2 adopts the updated 8-bit Adam optimizer in bitsandbytes\footnote{https://github.com/bitsandbytes-foundation/bitsandbytes}, which provides an option to directly generate low-rank updates.

\begin{figure}[H]
    \centering
    \includegraphics[width=\textwidth]{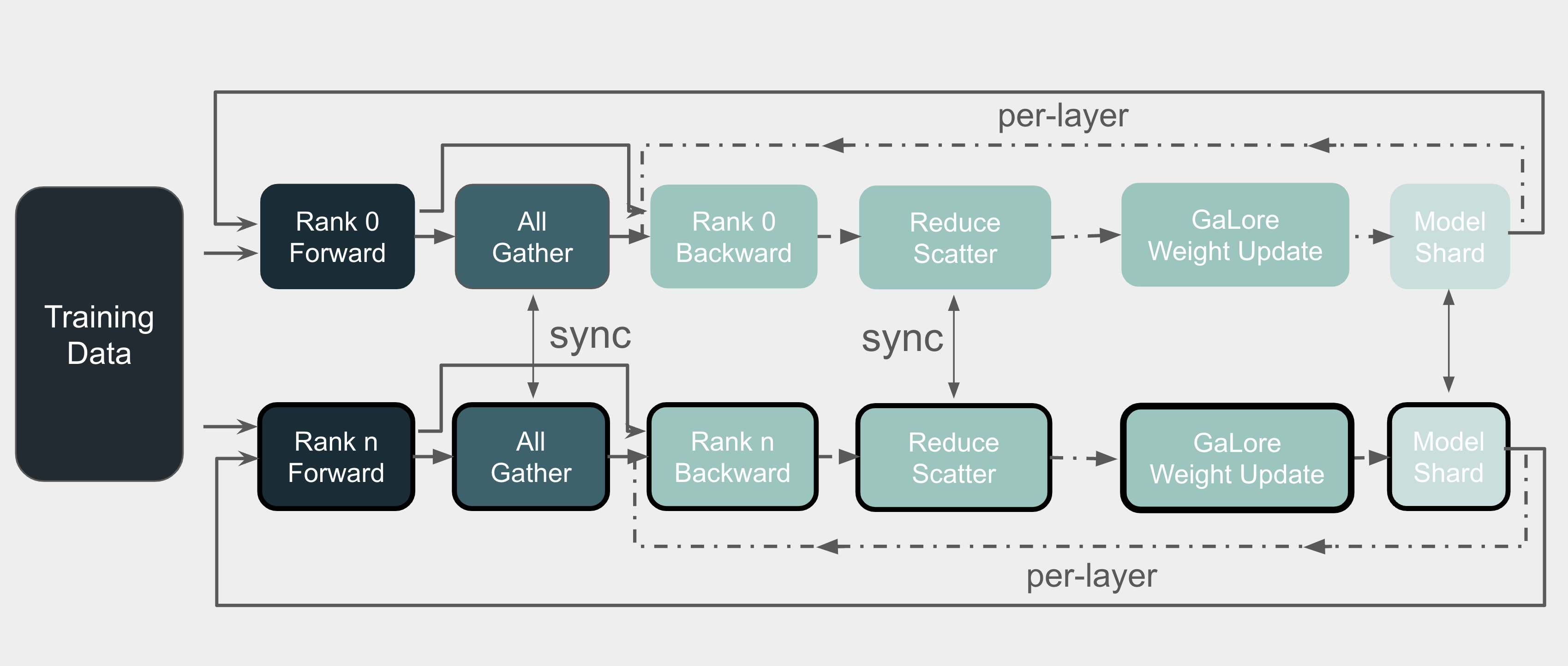} % Adjust the width as needed
    \captionsetup{justification=centering} % Center the caption
    \caption{Fully Sharded Data Parallelism.}
    \label{fig:FSDP}
\end{figure}

\subsection{GaLore 2 with FSDP}

GaLore 2 integrates with Fully Sharded Data Parallel (FSDP), a state-of-the-art training parallelization strategy. FSDP (Fig.~\ref{fig:FSDP}) is a sharding strategy that allows for efficient distributed training of large models across multiple GPUs. For the reader's reference, we provide a high-level overview of FSDP in the Appendix. 

Figure~\ref{fig:FSDP} illustrates the integration of GaLore with FSDP. FSDP introduces a new PyTorch hook that enables per-layer weight updates, fusing the backward pass and weight update to minimize gradient memory usage. Specifically, after reduce-scattering the gradient for each layer via FSDP, GaLore and its associated optimizer are called to update the weights. The gradient is discarded once the weight update is finished.
FSDP also provides other support for integrating GaLore, such as replicating SVD results across devices.

We compare the memory usage of GaLore and the baseline when FSDP is enabled. The evaluation is conducted on Llama 3 models with different model sizes during pre-training on the C4 dataset using the Adam optimizer. GaLore uses a quarter of full rank across all model sizes, and memory usage is measured in a 2-GPU setup with a sequence length of 2048 and a single batch size. As shown in Table~\ref{tab:memory_comparison}, GaLore demonstrates lower memory consumption compared to the baseline FSDP.

\begin{table}[h]
    \centering
    \begin{tabular}{lccc}
    \toprule
    \textbf{Model} & \textbf{Seq Length} &\textbf{Method} & \textbf{Memory per GPU} \\
    \midrule
    Llama3 8B & 4096 & GaLore + FSDP & 77.45GB \\
    Llama3 8B & 4096 & AdamW + FSDP & / \\
    Llama3 8B & 2048 & GaLore + FSDP & 72.84GB \\
    Llama3 8B & 2048 & AdamW + FSDP & 77.64GB \\
    \bottomrule
    \end{tabular}
    \caption{\centering Memory usage comparison per GPU for Llama3 models using FSDP.}
    \label{tab:memory_comparison}
\end{table}

\section{Scaling up GaLore 2 to 500 Billion Training Tokens}

In this section, we conducted large-scale experimentation on pre-training both GaLore and the baseline (8-bit Adam, proposed by \citet{dettmers8bitOptimizersBlockwise2021}) on 500 billion training tokens. 

We present the details of the Llama architecture and the hyperparameters used during pre-training. Table \ref{table:param} outlines the primary hyperparameters for Llama models across various sizes. All models utilize a maximum sequence length of 1024, with a total batch size of 1,048,576 tokens. For every experiment, we implement a learning rate warmup over the initial 10\% of training steps and employ cosine annealing for the learning rate schedule, reducing it to 10\% of its initial value.

We choose the rank as 1024 and tune the GaLore scale factor $\alpha$ from \{0.125, 0.250, 0.750, 0.1\} for the first 10B tokens to observe its training curve and select the one that leads to the most stable convergence behavior. Then, we finalize $\alpha=0.125$ for the remainder of our runs.
For each model, we used the optimal learning rate from the set \{0.01, 0.005, 0.001, 0.0005, 0.0001\}, selecting the best rate based on validation perplexity.

Similar to \cite{zhao2024galore}, we observe that GaLore is robust to hyperparameter variations and remains stable with a consistent learning rate across different model sizes. GaLore employs uniform hyperparameters for all models, including a subspace change frequency $T$ of 500. Notably, since $\alpha$ acts as a fractional learning rate, most modules (e.g., multi-head attention and feed-forward layers) in Llama models effectively use a learning rate of 0.000625. This remains a relatively large and stable learning rate compared to the full-rank baseline, which typically requires a learning rate $\leq 0.001$ to prevent training loss spikes.

\begin{table}[h]
\centering
\begin{tabular}{c c c c c}
\toprule
Params & Hidden & Intermediate & Heads & Layers \\
\midrule
7 B    & 4096   & 11008        & 32    & 32     \\
\bottomrule
\end{tabular}
\caption{\centering Hyperparameters for the Llama 7B model.}
\label{table:param}
\end{table}

We ran large-scale pretraining on 32 nodes of an H100 cluster. Each node has 8 $\times$ H100 (80GB) GPUs, for a total of 256 GPUs. We noticed that the H100 provides a speedup compared to running experiments on the A100, but the numerical effects on the end-to-end training results appear to be the same. 

To further analyze the performance, we conducted a detailed comparison of the validation loss curves for both GaLore and the baseline throughout the training process. The validation set, carefully curated to ensure no overlap with the training data, provides an unbiased measure of model generalization. 

\begin{figure}[h]
    \centering
    \includegraphics[width=0.7\textwidth]{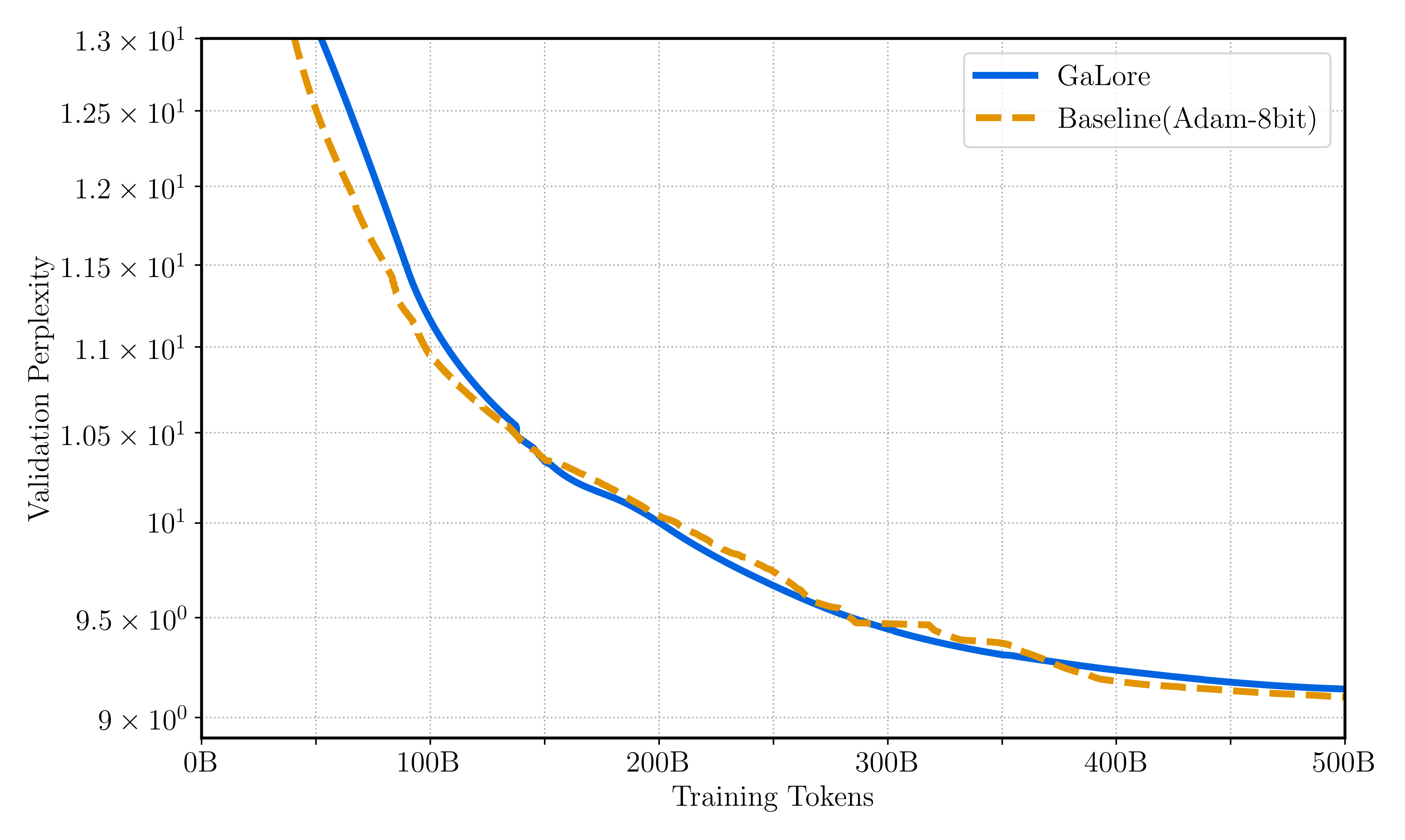} % Adjust the width as needed
    \captionsetup{justification=centering} % Center the caption
    \caption{Comparison of GaLore and Adam 8-bit baseline on the unseen validation set.}
    \label{fig:val_loss}
\end{figure}

As depicted in Figure \ref{fig:val_loss}, the validation loss for both methods exhibits a consistent downward trend, indicating effective learning.
Initially, during the first 150 billion tokens, GaLore's performance slightly lags behind the baseline. This phase can be attributed to the model's exploration of the optimization landscape, where GaLore's unique hyperparameter settings may require additional iterations to stabilize. However, as training progresses, GaLore begins to leverage its robust hyperparameter configuration, surpassing the baseline's performance around the 200 billion token mark.

Between 200 billion and 380 billion tokens, GaLore maintains a lower validation loss compared to the baseline, suggesting that it has found a more efficient optimization path. This period highlights GaLore's ability to adapt and optimize effectively.
Interestingly, around the 380 billion token mark, the baseline briefly overtakes GaLore. This fluctuation could be due to the inherent stochastic nature of the training process or temporary shifts in the optimization landscape. Despite this, GaLore quickly recovers and aligns closely with the baseline's performance as training approaches the 500 billion token mark.
By the end of the training, both GaLore and the baseline achieve comparable validation losses and perplexities, demonstrating GaLore's capability to match the baseline's performance.

This experiment underscores GaLore's capability in large-scale language model pre-training.

\begin{figure}[h]
    \centering
    \includegraphics[width=0.65\textwidth]{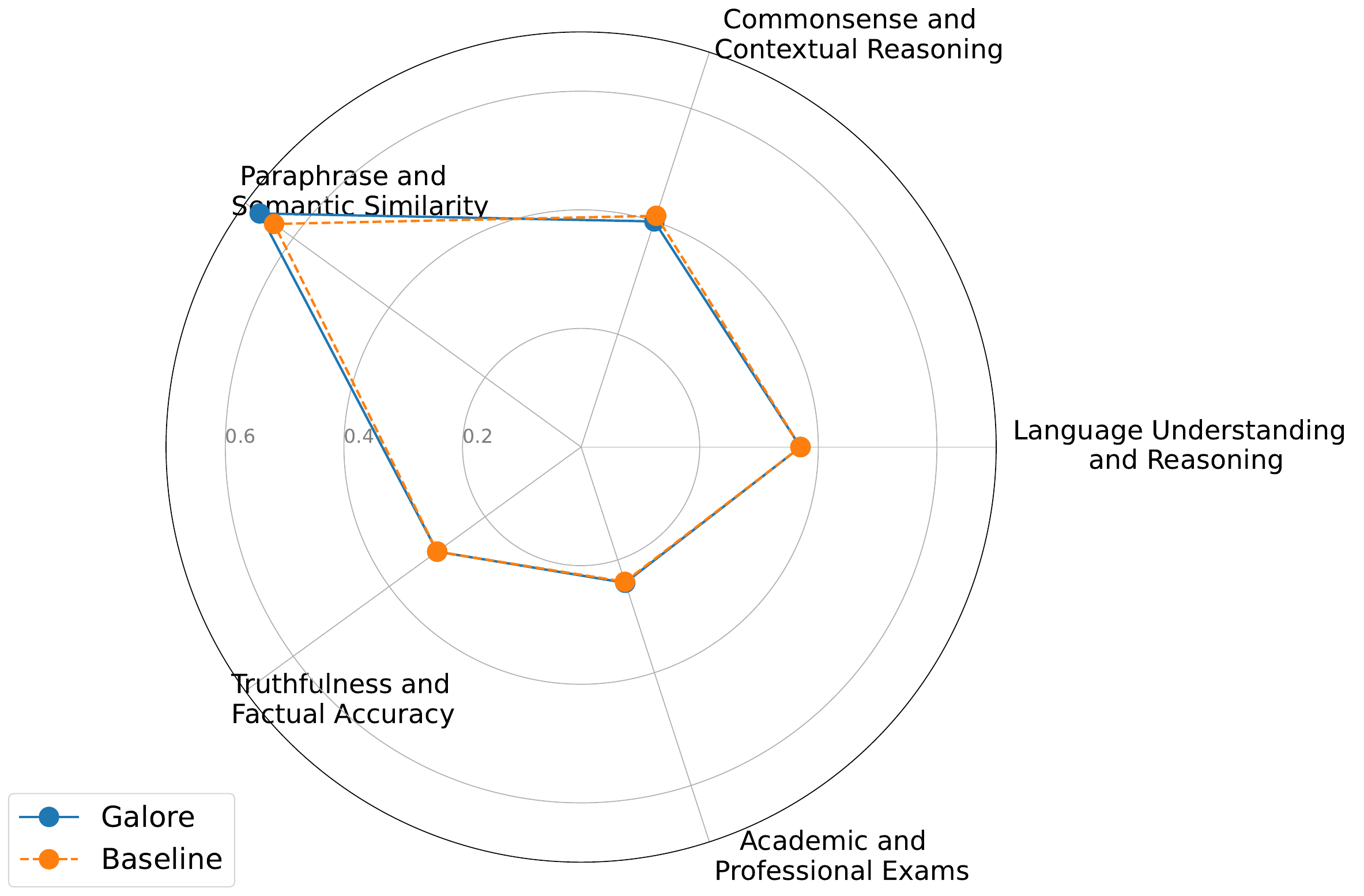} % Adjust the width as needed
    \captionsetup{justification=centering} % Center the caption
    \caption{Comparison of models across different categories}
    \label{fig:radar_plot}
\end{figure}

\begin{table}[h]
\centering
\begin{tabular}{lcc}
\toprule
\textbf{Language Understanding and Reasoning} & \textbf{Galore} & \textbf{Baseline} \\
\midrule
agieval\_en & 0.17 & 0.16 \\
agieval\_aqua\_rat & 0.27 & 0.24 \\
agieval\_gaokao\_english & 0.26 & 0.23 \\
agieval\_sat\_en & 0.28 & 0.28 \\
agieval\_sat\_en\_without\_passage & 0.29 & 0.26 \\
boolq & 0.56 & 0.56 \\
lambada\_openai & 0.31 & 0.31 \\
mnli & 0.33 & 0.33 \\
mnli\_mismatch & 0.33 & 0.32 \\
qnli & 0.49 & 0.52 \\
rte & 0.51 & 0.52 \\
sst2 & 0.51 & 0.51 \\
wnli & 0.52 & 0.45 \\
\midrule
\textbf{Average} & \textbf{0.37} & \textbf{0.37} \\
\bottomrule
\end{tabular}
\caption{Performance comparison of Galore and Baseline models on Language Understanding and Reasoning tasks.}
\label{tab:language_understanding}
\end{table}

\begin{table}[h]
\centering
\begin{tabular}{lcc}
\toprule
\textbf{Commonsense and Contextual Reasoning} & \textbf{Galore} & \textbf{Baseline} \\
\midrule
arc\_challenge & 0.27 & 0.28 \\
arc\_easy & 0.52 & 0.54 \\
hellaswag & 0.49 & 0.49 \\
ja\_leaderboard\_jcommonsenseqa & 0.19 & 0.19 \\
winogrande & 0.54 & 0.55 \\
\midrule
\textbf{Average} & \textbf{0.40} & \textbf{0.41} \\
\bottomrule
\end{tabular}
\caption{Performance comparison of Galore and Baseline models on Commonsense and Contextual Reasoning tasks.}
\label{tab:commonsense_reasoning}
\end{table}

\begin{table}[h]
\centering
\begin{tabular}{lcc}
\toprule
\textbf{Paraphrase and Semantic Similarity} & \textbf{Galore} & \textbf{Baseline} \\
\midrule
mrpc & 0.81 & 0.80 \\
qqp & 0.53 & 0.48 \\
\midrule
\textbf{Average} & \textbf{0.67} & \textbf{0.64} \\
\bottomrule
\end{tabular}
\caption{Performance comparison of Galore and Baseline models on Paraphrase and Semantic Similarity tasks.}
\label{tab:paraphrase_similarity}
\end{table}

\begin{table}[h]
\centering
\begin{tabular}{lcc}
\toprule
\textbf{Truthfulness and Factual Accuracy} & \textbf{Galore} & \textbf{Baseline} \\
\midrule
truthfulqa\_gen (bleu\_acc) & 0.31 & 0.31 \\
truthfulqa\_mc1 & 0.22 & 0.22 \\
truthfulqa\_mc2 & 0.38 & 0.38 \\
\midrule
\textbf{Average} & \textbf{0.30} & \textbf{0.30} \\
\bottomrule
\end{tabular}
\caption{Performance comparison of Galore and Baseline models on Truthfulness and Factual Accuracy.}
\label{tab:truthfulness}
\end{table}

\section{Downstream Performance}

In this section, we evaluate the performance of pretrained model checkpoints across various downstream tasks. We employed five-shot demonstrations, where applicable, to calculate the performance metrics. These metrics are categorized into five major groups: Language Understanding and Reasoning, Commonsense and Contextual Reasoning, Paraphrase and Semantic Similarity, Truthfulness and Factual Accuracy, and Academic and Professional Exams. It is important to note that the number of benchmarks within each category is not uniform.

This downstream evaluation performance metrics could provide insights into how each model behave in different areas of specialization.  
Our main table of results are shown in Table \ref{tab:language_understanding} to Table \ref{tab:academic_exams}.

\subsection{Language Understanding and Reasoning}

The evaluation of pretrained model checkpoints across diverse categories provides valuable insights into their capabilities. In the Language Understanding and Reasoning category, both models exhibit comparable performance, with an average score of 0.37. This suggests that they are equally proficient in tasks requiring language comprehension and reasoning, such as those found in the MNLI and QNLI benchmarks. The parity in performance indicates that both models have been effectively pretrained to handle a variety of linguistic constructs. Despite GaLore's low-rank optimization approach, it effectively captures the nuances of language, maintaining performance levels similar to the baseline. This demonstrates GaLore's ability to balance memory efficiency with robust language understanding capabilities.

\subsection{Commonsense and Contextual Reasoning}

In the Commonsense and Contextual Reasoning category, the baseline model slightly outperforms GaLore, with an average score of 0.41 compared to 0.40. A closer examination of the individual datasets within this category reveals that the baseline model's performance edge is minimal.

 On the ARC-Challenge dataset, the baseline model achieves a score of 0.28, which is just 0.01 points higher than GaLore's score. Similarly, on the ARC-Easy dataset, the baseline model's score of 0.54 is only 0.02 points above GaLore's performance. These marginal differences suggest that both models have comparable abilities in tackling the scientific reasoning and problem-solving tasks presented in the ARC datasets.

Similarly, the baseline model also slightly outperforms GaLore on the Winogrande dataset, with scores of 0.55 and 0.54, respectively. This dataset focuses on evaluating a model's ability to resolve ambiguous pronouns using commonsense reasoning. The narrow gap in performance indicates that both models are nearly equally proficient in this aspect of contextual understanding.

On the other hand,  when it comes to the HellaSwag and CommonsenseQA 2.0 datasets, the performance of the baseline model and GaLore is remarkably similar. These datasets assess a model's ability to draw upon commonsense knowledge to complete sentences or answer questions. The comparable scores suggest that both models are well-equipped to handle tasks that require a general understanding of real-world concepts and relationships.

\subsection{Paraphrase and Semantic Similarity}

In the Paraphrase and Semantic Similarity category, GaLore demonstrates a notable improvement over the baseline, with an average score of 0.67 compared to 0.64. 

GaLore's approach proves to be particularly effective in tasks that require a deep understanding of semantic nuances. This is evident from its performance on two key benchmarks in this category: the Microsoft Research Paraphrase Corpus (MRPC) and the Quora Question Pairs (QQP) dataset. 

On the MRPC benchmark, which tests a model's ability to identify paraphrases, GaLore achieves a score of 0.81, slightly higher than the baseline model's score of 0.80. This suggests that GaLore is better at recognizing when two sentences convey the same meaning, even if they are phrased differently. 

GaLore's performance on the QQP dataset is even more impressive, with a score of 0.53 compared to the baseline model's 0.48, achieving a +0.5 points improvement. The QQP dataset consists of pairs of questions from Quora, a popular question-answering platform, and the task is to determine whether the questions are semantically equivalent. GaLore's higher score indicates that it is more adept at understanding the semantic relationships between questions and identifying those that are asking the same thing.

\subsection{Truthfulness and Factual Accuracy}

In the TruthfulQA generation benchmark (truthfulqa\_gen), which measures the accuracy of the generated content using BLEU scores, both models achieve a score of 0.31. This indicates that GaLore and the baseline model are equally proficient at generating truthful and factually accurate content.
Similarly, in the TruthfulQA multiple-choice benchmarks (truthfulqa\_mc1 and truthfulqa\_mc2), both models perform identically, with scores of 0.22 and 0.38, respectively. These benchmarks assess the models' ability to select the most truthful and factually accurate answer from a set of options, further confirming their parity in this category.

The identical performance of GaLore and the baseline model in the Truthfulness and Factual Accuracy category suggests that GaLore's low-rank optimization approach does not compromise its ability to maintain factual accuracy and truthfulness in generated content. This is a crucial finding, as it demonstrates that the memory efficiency benefits of GaLore's approach do not come at the cost of decreased performance in this important aspect of language generation.

\begin{table}[H]
\centering
\begin{tabular}{lcc}
\toprule
\textbf{Academic and Professional Exams} & \textbf{Galore} & \textbf{Baseline} \\
\midrule
agieval\_logiqa\_en & 0.27 & 0.26 \\
agieval\_lsat\_ar & 0.24 & 0.26 \\
agieval\_lsat\_lr & 0.22 & 0.22 \\
agieval\_lsat\_rc & 0.18 & 0.15 \\
% agieval\_math & 0.01 & 0.01 \\
agieval\_sat\_math & 0.24 & 0.23 \\
mmlu & 0.25 & 0.26 \\
mmlu\_humanities & 0.26 & 0.25 \\
mmlu\_other & 0.26 & 0.27 \\
mmlu\_social\_sciences & 0.24 & 0.24 \\
mmlu\_stem & 0.26 & 0.27 \\
\midrule
\textbf{Average} & \textbf{0.24} & \textbf{0.24} \\
\bottomrule
\end{tabular}
\caption{Performance comparison of Galore and Baseline models on Academic and Professional Exams.}
\label{tab:academic_exams}
\end{table}

\subsection{Academic and Professional Exams}

In the Academic and Professional Exams category, both models achieve the same average score of 0.24. This suggests that GaLore's memory-efficient approach does not detract from its ability to perform in academic and professional exam settings. The comparable performance indicates that both models are well-suited for tasks that require logical reasoning and problem-solving skills, as demonstrated in benchmarks like LSAT and SAT.

Overall, the downstream performance evaluation reveals that GaLore, with its low-rank optimization, maintains competitive performance across various categories while offering significant memory efficiency benefits. Although the baseline model slightly outperforms GaLore in commonsense reasoning tasks, GaLore excels in paraphrase and semantic similarity tasks, showcasing its potential as a memory-efficient alternative for large-scale language model training. The results underscore GaLore's capability to balance computational efficiency with robust performance, making it a promising approach for future advancements in language model training.

\section{Conclusion}
In this report, we introduced GaLore 2, an enhanced framework for memory-efficient training of large language models. GaLore 2 addresses key challenges in the original GaLore approach, such as the computational overhead of subspace updates and integration with advanced parallelization strategies like FSDP. By incorporating fast randomized SVD and supporting recent advancements like low-bit quantization and higher-order tensor structures, GaLore 2 offers a scalable and efficient solution for training large models.

Our large-scale experiments demonstrated GaLore's capability to pre-train a LLaMA-7B language model on 500 billion tokens, achieving comparable performance to baseline methods while significantly reducing memory consumption. The downstream evaluation further highlighted GaLore's competitive performance across various tasks, particularly excelling in paraphrase and semantic similarity tasks.

Overall, GaLore 2 represents a significant step forward in making large-scale language model training more accessible and efficient, without compromising on performance. Its ability to balance computational efficiency with robust model capabilities makes it a promising approach for future advancements in the field of natural language processing.

\section{Acknowledgments}
We would like to thank the following individuals for their contributions and discussions that have significantly improved this work: David Pitt, Jean Kossaifi, Joel Tropp, Mark Saroufim, Mark Tygert, Robert Joseph, Wei Feng, and Zhengyu Zhang.

\clearpage
\newpage
\bibliographystyle{assets/plainnat}
\bibliography{paper}

\begin{thebibliography}{38}
\providecommand{\natexlab}[1]{#1}
\providecommand{\url}[1]{\texttt{#1}}
\expandafter\ifx\csname urlstyle\endcsname\relax
  \providecommand{\doi}[1]{doi: #1}\else
  \providecommand{\doi}{doi: \begingroup \urlstyle{rm}\Url}\fi

\bibitem[Brown et~al.(2020)Brown, Mann, Ryder, Subbiah, Kaplan, Dhariwal, Neelakantan, Shyam, Sastry, Askell, et~al.]{ll3}
Tom Brown, Benjamin Mann, Nick Ryder, Melanie Subbiah, Jared~D Kaplan, Prafulla Dhariwal, Arvind Neelakantan, Pranav Shyam, Girish Sastry, Amanda Askell, et~al.
\newblock Language models are few-shot learners.
\newblock \emph{Advances in neural information processing systems}, 33:\penalty0 1877--1901, 2020.

\bibitem[Cohen et~al.(2025)Cohen, Gromov, Yang, and Tian]{cohen2025spectral}
Andrew Cohen, Andrey Gromov, Kaiyu Yang, and Yuandong Tian.
\newblock Spectral journey: How transformers predict the shortest path.
\newblock \emph{arXiv preprint arXiv:2502.08794}, 2025.

\bibitem[Das(2024)]{naturalgalore}
Arijit Das.
\newblock Natural galore: Accelerating galore for memory-efficient llm training and fine-tuning.
\newblock \emph{ArXiv}, abs/2410.16029, 2024.
\newblock \url{https://api.semanticscholar.org/CorpusID:273501786}.

\bibitem[Dettmers et~al.(2022)Dettmers, Lewis, Shleifer, and Zettlemoyer]{dettmers8bitOptimizersBlockwise2021}
Tim Dettmers, Mike Lewis, Sam Shleifer, and Luke Zettlemoyer.
\newblock 8-bit optimizers via block-wise quantization.
\newblock In \emph{The Tenth International Conference on Learning Representations, {ICLR} 2022, Virtual Event, April 25-29, 2022}. OpenReview.net, 2022.

\bibitem[Dubey et~al.(2024)Dubey, Jauhri, Pandey, Kadian, Al-Dahle, Letman, Mathur, Schelten, Yang, Fan, et~al.]{ll1}
Abhimanyu Dubey, Abhinav Jauhri, Abhinav Pandey, Abhishek Kadian, Ahmad Al-Dahle, Aiesha Letman, Akhil Mathur, Alan Schelten, Amy Yang, Angela Fan, et~al.
\newblock The llama 3 herd of models.
\newblock \emph{arXiv preprint arXiv:2407.21783}, 2024.

\bibitem[George et~al.(2024)George, Pitt, Zhao, Kossaifi, Luo, Tian, and Anandkumar]{george2024tensorgalore}
Robert~Joseph George, David Pitt, Jiawei Zhao, Jean Kossaifi, Cheng Luo, Yuandong Tian, and Anima Anandkumar.
\newblock Tensor-galore: Memory-efficient training via gradient tensor decomposition.
\newblock In \emph{OPT 2024: Optimization for Machine Learning}, 2024.
\newblock \url{https://openreview.net/forum?id=sBaUZzZXJN}.

\bibitem[Halko et~al.(2011)Halko, Martinsson, and Tropp]{halko2011finding}
Nathan Halko, Per-Gunnar Martinsson, and Joel~A Tropp.
\newblock Finding structure with randomness: Probabilistic algorithms for constructing approximate matrix decompositions.
\newblock \emph{SIAM review}, 53\penalty0 (2):\penalty0 217--288, 2011.

\bibitem[Hao et~al.(2024)Hao, Sukhbaatar, Su, Li, Hu, Weston, and Tian]{ds3}
Shibo Hao, Sainbayar Sukhbaatar, DiJia Su, Xian Li, Zhiting Hu, Jason Weston, and Yuandong Tian.
\newblock Training large language models to reason in a continuous latent space.
\newblock \emph{arXiv preprint arXiv:2412.06769}, 2024.

\bibitem[Hu et~al.(2021)Hu, Shen, Wallis, Allen-Zhu, Li, Wang, Wang, and Chen]{hu2021lora}
Edward~J. Hu, Yelong Shen, Phillip Wallis, Zeyuan Allen-Zhu, Yuanzhi Li, Shean Wang, Lu~Wang, and Weizhu Chen.
\newblock Lora: Low-rank adaptation of large language models, 2021.
\newblock \url{https://arxiv.org/abs/2106.09685}.

\bibitem[Jaiswal et~al.(2024)Jaiswal, Yin, Zhang, Liu, Zhao, Tian, and Wang]{welore}
Ajay~Kumar Jaiswal, Lu~Yin, Zhenyu~(Allen) Zhang, Shiwei Liu, Jiawei Zhao, Yuandong Tian, and Zhangyang Wang.
\newblock From galore to welore: How low-rank weights non-uniformly emerge from low-rank gradients.
\newblock \emph{ArXiv}, abs/2407.11239, 2024.
\newblock \url{https://api.semanticscholar.org/CorpusID:271218569}.

\bibitem[Kossaifi et~al.(2019)Kossaifi, Panagakis, Anandkumar, and Pantic]{tensorly}
Jean Kossaifi, Yannis Panagakis, Anima Anandkumar, and Maja Pantic.
\newblock Tensorly: Tensor learning in python.
\newblock \emph{Journal of Machine Learning Research}, 20\penalty0 (26):\penalty0 1--6, 2019.
\newblock \url{http://jmlr.org/papers/v20/18-277.html}.

\bibitem[Lehnert et~al.(2024)Lehnert, Sukhbaatar, Su, Zheng, Mcvay, Rabbat, and Tian]{ds2}
Lucas Lehnert, Sainbayar Sukhbaatar, DiJia Su, Qinqing Zheng, Paul Mcvay, Michael Rabbat, and Yuandong Tian.
\newblock Beyond a*: Better planning with transformers via search dynamics bootstrapping.
\newblock \emph{arXiv preprint arXiv:2402.14083}, 2024.

\bibitem[Lialin et~al.(2023)Lialin, Muckatira, Shivagunde, and Rumshisky]{lialin2023relora}
Vladislav Lialin, Sherin Muckatira, Namrata Shivagunde, and Anna Rumshisky.
\newblock Relora: High-rank training through low-rank updates.
\newblock In \emph{Workshop on Advancing Neural Network Training: Computational Efficiency, Scalability, and Resource Optimization (WANT@ NeurIPS 2023)}, 2023.

\bibitem[Liang et~al.(2024)Liang, Liu, Chen, and Liu]{liang2024memoryefficientllmtrainingonline}
Kaizhao Liang, Bo~Liu, Lizhang Chen, and Qiang Liu.
\newblock Memory-efficient llm training with online subspace descent, 2024.
\newblock \url{https://arxiv.org/abs/2408.12857}.

\bibitem[Lin et~al.(2025)Lin, Jin, Xu, Wu, Sukhbaatar, Zhu, He, Chen, Weston, Tian, et~al.]{lin2025step}
Yen-Ting Lin, Di~Jin, Tengyu Xu, Tianhao Wu, Sainbayar Sukhbaatar, Chen Zhu, Yun He, Yun-Nung Chen, Jason Weston, Yuandong Tian, et~al.
\newblock Step-kto: Optimizing mathematical reasoning through stepwise binary feedback.
\newblock \emph{arXiv preprint arXiv:2501.10799}, 2025.

\bibitem[Liu et~al.(2024{\natexlab{a}})Liu, Zhao, Fedorov, Soran, Choudhary, Krishnamoorthi, Chandra, Tian, and Blankevoort]{liu2024spinquant}
Zechun Liu, Changsheng Zhao, Igor Fedorov, Bilge Soran, Dhruv Choudhary, Raghuraman Krishnamoorthi, Vikas Chandra, Yuandong Tian, and Tijmen Blankevoort.
\newblock Spinquant: Llm quantization with learned rotations.
\newblock \emph{arXiv preprint arXiv:2405.16406}, 2024{\natexlab{a}}.

\bibitem[Liu et~al.(2024{\natexlab{b}})Liu, Zhao, Fedorov, Soran, Choudhary, Krishnamoorthi, Chandra, Tian, and Blankevoort]{liu2024spinquantllmquantizationlearned}
Zechun Liu, Changsheng Zhao, Igor Fedorov, Bilge Soran, Dhruv Choudhary, Raghuraman Krishnamoorthi, Vikas Chandra, Yuandong Tian, and Tijmen Blankevoort.
\newblock Spinquant: Llm quantization with learned rotations.
\newblock 2024{\natexlab{b}}.
\newblock \url{https://arxiv.org/abs/2405.16406}.

\bibitem[Luo et~al.(2024)Luo, Yu, and Li]{luo2024badam}
Qijun Luo, Hengxu Yu, and Xiao Li.
\newblock Badam: A memory efficient full parameter training method for large language models.
\newblock \emph{arXiv preprint arXiv:2404.02827}, 2024.

\bibitem[Lv et~al.(2023)Lv, Yan, Guo, Lv, and Qiu]{lvAdaLomoLowmemoryOptimization2023}
Kai Lv, Hang Yan, Qipeng Guo, Haijun Lv, and Xipeng Qiu.
\newblock {{AdaLomo}}: {{Low-memory Optimization}} with {{Adaptive Learning Rate}}.
\newblock \emph{ArXiv preprint arXiv:2310.10195}, 2023.

\bibitem[Pan et~al.(2024)Pan, Liu, Diao, Pi, Zhang, Han, and Zhang]{pan2024lisa}
Rui Pan, Xiang Liu, Shizhe Diao, Renjie Pi, Jipeng Zhang, Chi Han, and Tong Zhang.
\newblock Lisa: Layerwise importance sampling for memory-efficient large language model fine-tuning.
\newblock \emph{arXiv preprint arXiv:2403.17919}, 2024.

\bibitem[Paulus et~al.(2024)Paulus, Zharmagambetov, Guo, Amos, and Tian]{paulus2024advprompter}
Anselm Paulus, Arman Zharmagambetov, Chuan Guo, Brandon Amos, and Yuandong Tian.
\newblock Advprompter: Fast adaptive adversarial prompting for llms.
\newblock \emph{arXiv preprint arXiv:2404.16873}, 2024.

\bibitem[Pedregosa et~al.(2018)Pedregosa, Varoquaux, Gramfort, Michel, Thirion, Grisel, Blondel, Müller, Nothman, Louppe, Prettenhofer, Weiss, Dubourg, Vanderplas, Passos, Cournapeau, Brucher, Perrot, and Édouard Duchesnay]{scikitlearn}
Fabian Pedregosa, Gaël Varoquaux, Alexandre Gramfort, Vincent Michel, Bertrand Thirion, Olivier Grisel, Mathieu Blondel, Andreas Müller, Joel Nothman, Gilles Louppe, Peter Prettenhofer, Ron Weiss, Vincent Dubourg, Jake Vanderplas, Alexandre Passos, David Cournapeau, Matthieu Brucher, Matthieu Perrot, and Édouard Duchesnay.
\newblock Scikit-learn: Machine learning in python, 2018.
\newblock \url{https://arxiv.org/abs/1201.0490}.

\bibitem[Robert et~al.(2024)Robert, Safaryan, Modoranu, and Alistarh]{Robert2024LDAdamAO}
Thomas Robert, M.~H. Safaryan, Ionut-Vlad Modoranu, and Dan Alistarh.
\newblock Ldadam: Adaptive optimization from low-dimensional gradient statistics.
\newblock \emph{ArXiv}, abs/2410.16103, 2024.
\newblock \url{https://api.semanticscholar.org/CorpusID:273502249}.

\bibitem[Shazeer and Stern(2018)]{shazeerAdafactorAdaptiveLearning}
Noam Shazeer and Mitchell Stern.
\newblock Adafactor: Adaptive learning rates with sublinear memory cost.
\newblock In \emph{Proceedings of the 35th International Conference on Machine Learning, {ICML} 2018, Stockholmsm{\"{a}}ssan, Stockholm, Sweden, July 10-15, 2018}. {PMLR}, 2018.

\bibitem[Su et~al.(2020)Su, Ooi, Lu, Schuurmans, and Boutilier]{su2020conqur}
Dijia Su, Jayden Ooi, Tyler Lu, Dale Schuurmans, and Craig Boutilier.
\newblock Conqur: Mitigating delusional bias in deep q-learning.
\newblock In \emph{International Conference on Machine Learning}, pages 9187--9195. PMLR, 2020.

\bibitem[Su et~al.(2021)Su, Lee, Mulvey, and Poor]{su2021musbo}
DiJia Su, Jason~D Lee, John~M Mulvey, and H~Vincent Poor.
\newblock Musbo: Model-based uncertainty regularized and sample efficient batch optimization for deployment constrained reinforcement learning.
\newblock \emph{arXiv preprint arXiv:2102.11448}, 2021.

\bibitem[Su et~al.(2024)Su, Sukhbaatar, Rabbat, Tian, and Zheng]{ds1}
DiJia Su, Sainbayar Sukhbaatar, Michael Rabbat, Yuandong Tian, and Qinqing Zheng.
\newblock Dualformer: Controllable fast and slow thinking by learning with randomized reasoning traces.
\newblock \emph{arXiv preprint arXiv:2410.09918}, 2024.

\bibitem[Su et~al.(2025)Su, Zhu, Xu, Jiao, Tian, and Zheng]{su2025token}
DiJia Su, Hanlin Zhu, Yingchen Xu, Jiantao Jiao, Yuandong Tian, and Qinqing Zheng.
\newblock Token assorted: Mixing latent and text tokens for improved language model reasoning.
\newblock \emph{arXiv preprint arXiv:2502.03275}, 2025.

\bibitem[Su et~al.(2022)Su, Douillard, Al-Rfou, Park, and Sapp]{su2022narrowing}
DiJia~Andy Su, Bertrand Douillard, Rami Al-Rfou, Cheol Park, and Benjamin Sapp.
\newblock Narrowing the coordinate-frame gap in behavior prediction models: Distillation for efficient and accurate scene-centric motion forecasting.
\newblock In \emph{2022 International Conference on Robotics and Automation (ICRA)}, pages 653--659. IEEE, 2022.

\bibitem[Touvron et~al.(2023)Touvron, Lavril, Izacard, Martinet, Lachaux, Lacroix, Rozi{\`e}re, Goyal, Hambro, Azhar, et~al.]{ll2}
Hugo Touvron, Thibaut Lavril, Gautier Izacard, Xavier Martinet, Marie-Anne Lachaux, Timoth{\'e}e Lacroix, Baptiste Rozi{\`e}re, Naman Goyal, Eric Hambro, Faisal Azhar, et~al.
\newblock Llama: Open and efficient foundation language models.
\newblock \emph{arXiv preprint arXiv:2302.13971}, 2023.

\bibitem[Wang et~al.(2024)Wang, Yang, Zhu, Yang, Cohen, Li, and Tian]{ww1}
Danqing Wang, Kevin Yang, Hanlin Zhu, Xiaomeng Yang, Andrew Cohen, Lei Li, and Yuandong Tian.
\newblock Learning personalized alignment for evaluating open-ended text generation.
\newblock In Yaser Al-Onaizan, Mohit Bansal, and Yun-Nung Chen, editors, \emph{Proceedings of the 2024 Conference on Empirical Methods in Natural Language Processing}, pages 13274--13292, Miami, Florida, USA, November 2024. Association for Computational Linguistics.
\newblock \doi{10.18653/v1/2024.emnlp-main.737}.
\newblock \url{https://aclanthology.org/2024.emnlp-main.737}.

\bibitem[Wang et~al.()Wang, Caccia, Ostapenko, Yuan, and Sordoni]{wang2310guiding}
Xinyi Wang, Lucas Caccia, Oleksiy Ostapenko, Xingdi Yuan, and Alessandro Sordoni.
\newblock Guiding language model reasoning with planning tokens, december 2023b.
\newblock \emph{URL http://arxiv. org/abs/2310.05707}.

\bibitem[Wang et~al.(2023)Wang, Lin, Zeng, and Zhang]{wang2023multilora}
Yiming Wang, Yu~Lin, Xiaodong Zeng, and Guannan Zhang.
\newblock Multilora: Democratizing lora for better multi-task learning.
\newblock \emph{arXiv preprint arXiv:2311.11501}, 2023.

\bibitem[Wu et~al.(2024)Wu, Yuan, Golovneva, Xu, Tian, Jiao, Weston, and Sukhbaatar]{wu2024meta}
Tianhao Wu, Weizhe Yuan, Olga Golovneva, Jing Xu, Yuandong Tian, Jiantao Jiao, Jason Weston, and Sainbayar Sukhbaatar.
\newblock Meta-rewarding language models: Self-improving alignment with llm-as-a-meta-judge.
\newblock \emph{arXiv preprint arXiv:2407.19594}, 2024.

\bibitem[Zhang et~al.(2024)Zhang, Jaiswal, Yin, Liu, Zhao, Tian, and Wang]{zhang2024q}
Zhenyu Zhang, Ajay Jaiswal, Lu~Yin, Shiwei Liu, Jiawei Zhao, Yuandong Tian, and Zhangyang Wang.
\newblock Q-galore: Quantized galore with int4 projection and layer-adaptive low-rank gradients.
\newblock \emph{arXiv preprint arXiv:2407.08296}, 2024.

\bibitem[Zhao et~al.(2024)Zhao, Zhang, Chen, Wang, Anandkumar, and Tian]{zhao2024galore}
Jiawei Zhao, Zhenyu Zhang, Beidi Chen, Zhangyang Wang, Anima Anandkumar, and Yuandong Tian.
\newblock Galore: Memory-efficient llm training by gradient low-rank projection.
\newblock \emph{arXiv preprint arXiv:2403.03507}, 2024.

\bibitem[Zhou et~al.(2025{\natexlab{a}})Zhou, Liu, Chen, Tian, and Chen]{zhou2025gsm}
Yang Zhou, Hongyi Liu, Zhuoming Chen, Yuandong Tian, and Beidi Chen.
\newblock Gsm-infinite: How do your llms behave over infinitely increasing context length and reasoning complexity?
\newblock \emph{arXiv preprint arXiv:2502.05252}, 2025{\natexlab{a}}.

\bibitem[Zhou et~al.(2025{\natexlab{b}})Zhou, Jiang, Tian, Weston, Levine, Sukhbaatar, and Li]{zhou2025sweet}
Yifei Zhou, Song Jiang, Yuandong Tian, Jason Weston, Sergey Levine, Sainbayar Sukhbaatar, and Xian Li.
\newblock Sweet-rl: Training multi-turn llm agents on collaborative reasoning tasks.
\newblock \emph{arXiv preprint arXiv:2503.15478}, 2025{\natexlab{b}}.

\end{thebibliography}

\clearpage
\newpage
\beginappendix

\section{Applying GaLore on Adam optimizer}

\citet{zhao2024galore} provides an algorithm that applies GaLore to the Adam optimizer, as shown in Algorithm~\ref{alg:galore_adam}. It projects the gradient onto low-rank subspace before feeding into the Adam optimizer. The low-rank normalized gradient generated by Adam is reprojected back to full-rank parameter space before applying the update to the original weights. GaLore can be applied to other preconditioned optimizers in a similar way.

\begin{algorithm}[h]
    \caption{Adam with GaLore}
    \label{alg:galore_adam}
  \begin{algorithmic}
    \STATE {\bfseries Input:} A layer weight matrix $W \in \mathbb{R}^{m \times n}$ with $m \leq n$. Step size $\eta$, scale factor $\alpha$, decay rates $\beta_1, \beta_2$, rank $r$, subspace change frequency $T$.
    \STATE Initialize first-order moment $M_0 \in \mathbb{R}^{n \times r} \gets 0$
    \STATE Initialize second-order moment $V_0 \in \mathbb{R}^{n \times r} \gets 0$
    \STATE Initialize step $t \gets 0$
    \REPEAT
    \STATE $G_t \in \mathbb{R}^{m \times n} \gets - \nabla_W \phi_t(W_t)$ 
    \IF{$t \bmod T = 0$}
    \STATE $U, S, V \gets \text{SVD}(G_t)$
    \STATE $P_t \gets U[:, :r]$ \hfill \COMMENT{Initialize left projector as $m \leq n$}
    \ELSE
    \STATE $P_t \gets P_{t-1}$ \hfill \COMMENT{Reuse the previous projector}
    \ENDIF
    \STATE $R_t \gets P_{t}^{\top} G_t$ \hfill \COMMENT{Project gradient into compact space}
    \\\hrulefill
    % Sub-algorithm starts here
    \STATE {\bfseries $\update(R_t)$ by Adam}
    % \bindent
    \hspace{\algorithmicindent} \STATE $M_t \gets \beta_1 \cdot M_{t-1} + (1 - \beta_1) \cdot R_t$ 
    \hspace{\algorithmicindent} \STATE $V_t \gets \beta_2 \cdot V_{t-1} + (1 - \beta_2) \cdot R_t^2$ 
    \hspace{\algorithmicindent} \STATE $M_t \gets M_t / (1 - \beta_1^t)$
    \hspace{\algorithmicindent} \STATE $V_t \gets V_t / (1 - \beta_2^t)$ 
    \hspace{\algorithmicindent} \STATE $N_t \gets M_t / (\sqrt{V_t} + \epsilon)$
    % \eindent
    % Sub-algorithm ends here
    \\\hrulefill
    \STATE $\tilde G_t \gets \alpha \cdot P N_t$ \hfill \COMMENT{Project back to original space}
    \STATE $W_t \gets W_{t-1} + \eta \cdot \tilde G_t$
    \STATE $t \gets t + 1$
    \UNTIL{convergence criteria met}
    \RETURN $W_t$
  \end{algorithmic}
 \end{algorithm}

\section{Distributed Data Parallel (DDP)}
Distributed Data Parallel (DDP) is a widely adopted method for parallelizing the training of deep learning models across multiple GPUs or nodes. The primary objective of DDP is to distribute the data across different devices, allowing each device to process a subset of the data independently. This approach leverages data parallelism, where each GPU contains a replicate of the model and each processes a mini-batch of data and computes gradients locally. Then, the computed gradients are then averaged across (all-reduce) over all replicas, and the update the model synchronously. 

The key advantage of DDP is its simplicity and ease of integration into existing training pipelines, making it a preferred choice for many practitioners.

\begin{figure}[h]
    \centering
    \includegraphics[width=\textwidth]{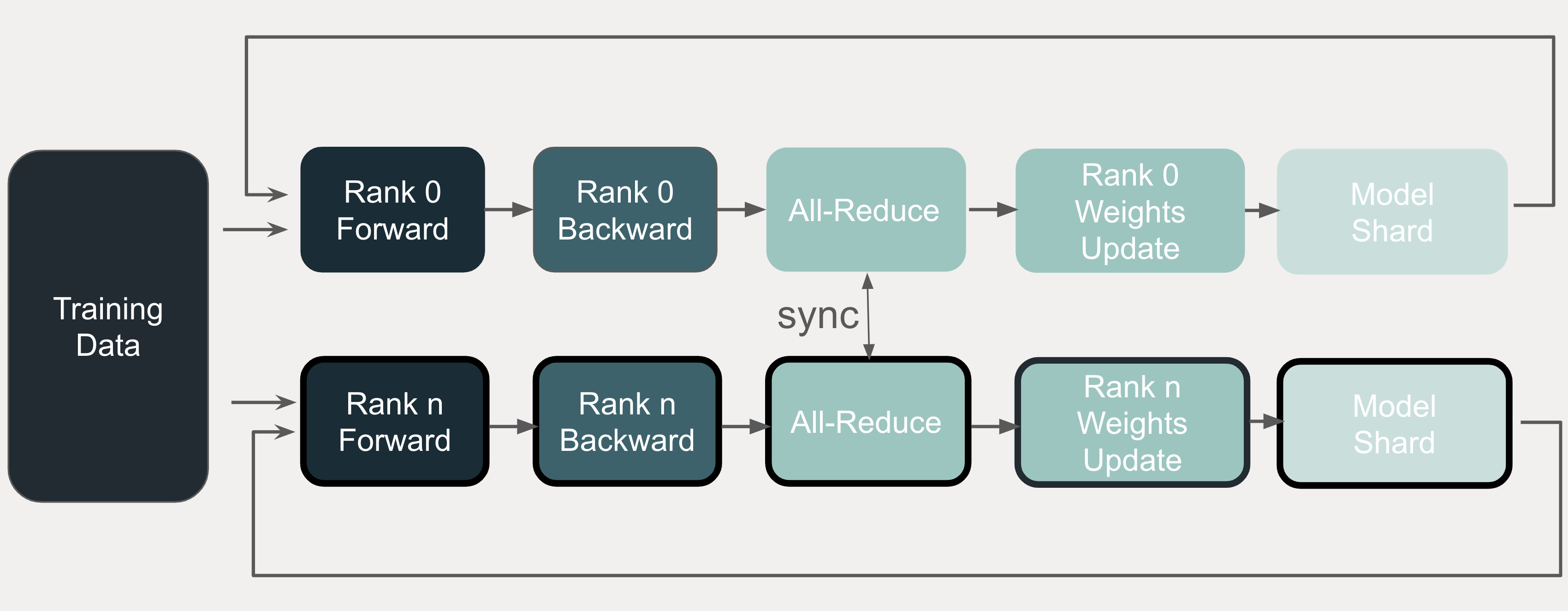} % Adjust the width as needed
    \captionsetup{justification=centering} % Center the caption
    \caption{Distributed Data Parallelism.}
    \label{fig:DDP}
\end{figure}

\section{Fully Sharded Data Parallel (FSDP) }
FSDP is an advanced parallelization technique designed to address the limitations of traditional data parallelism methods like DDP. FSDP introduces a more granular level of parallelism by sharding both the model parameters and optimizer states across multiple devices or nodes.  This approach significantly reduces memory consumption, enabling the training of larger models that would otherwise be infeasible with DDP.

FSDP operates by partitioning the model into smaller shards, each of which is distributed across different devices or nodes. During the forward and backward passes, only the necessary shards are loaded into memory, while the rest remain offloaded. This strategy not only optimizes memory usage but also reduces communication overhead, as only the relevant shards are synchronized during gradient updates.

The major advantage of FSDP over DDP is its memory efficiency.  As mentioned previously, DDP requires each device to maintain a full copy of the model parameters, leading to high memory consumption, especially for large models. For instance, modern size LLM on the scale of 7B to 400B parameter size might not be able to fit into one single GPU.  

In contrast, FSDP's sharding mechanism allows for a more efficient use of memory, enabling the training of models with significantly larger parameter counts.

DDP involves synchronizing gradients across all devices, which can introduce substantial communication overhead, particularly in distributed settings with limited bandwidth. FSDP mitigates this issue by reducing the amount of data that needs to be communicated, as only the relevant shards are synchronized. This results in improved scalability and performance, especially in large-scale distributed environments.
Complexity and Implementation
While DDP is relatively straightforward to implement and integrate into existing workflows, FSDP introduces additional complexity due to its sharding mechanism. The partitioning of model parameters and optimizer states requires careful management to ensure efficient training. However, recent advancements in deep learning frameworks have simplified the implementation of FSDP, making it more accessible to researchers and practitioners.

\end{document}